\documentclass[journal]{IEEEtran}

\usepackage{amscd,amssymb}
\usepackage{amsfonts}
\usepackage{graphicx,epsf}
\usepackage{enumerate,color}
\usepackage[mathcal]{euscript}
\usepackage[dvipsnames]{xcolor}
\usepackage{hyperref}
\usepackage{subfigure}
\usepackage{amsmath}
\usepackage{amssymb}
\usepackage{amsfonts}
\usepackage{dsfont}

\usepackage{algorithm}
\usepackage{algpseudocode}

\newcommand*\Let[2]{\State #1 $\gets$ #2}
\newcommand{\argminsub}[1]{\underset{{ #1 }}{{\rm argmin}}}

\newcommand{\sqnorm}[1]{\Vert #1 \Vert_2^2}
\newcommand{\R}{{\mathbb R}}

\newcommand{\revision}[1]{{ #1}}
\newcommand{\revisNew}[1]{{ #1}}

\def\kwave/{\textbf{k}-\texttt{Wave}}

\begin{document}
%
% paper title
% Titles are generally capitalized except for words such as a, an, and, as,
% at, but, by, for, in, nor, of, on, or, the, to and up, which are usually
% not capitalized unless they are the first or last word of the title.
% Linebreaks \\ can be used within to get better formatting as desired.
% Do not put math or special symbols in the title.
\title{Model based learning for accelerated, limited-view 3D photoacoustic tomography}
%
%
% author names and IEEE memberships
% note positions of commas and nonbreaking spaces ( ~ ) LaTeX will not break
% a structure at a ~ so this keeps an author's name from being broken across
% two lines.
% use \thanks{} to gain access to the first footnote area
% a separate \thanks must be used for each paragraph as LaTeX2e's \thanks
% was not built to handle multiple paragraphs
%

\author{ 
{\small Andreas Hauptmann, Felix Lucka, Marta Betcke, Nam Huynh, Jonas Adler, Ben Cox, Paul Beard, Sebastien Ourselin, and Simon Arridge}% 

\thanks{Andreas Hauptmann, Felix Lucka, Marta Betcke, Sebastien Ourselin, and Simon Arridge are with the Department of Computer Science, University College London, London, United Kingdom } %

\thanks{Nam Huynh, Ben Cox, and Paul Beard are with the {Department of Medical Physics and Biomedical Engineering, University College London, London, United Kingdom }
}

\thanks{Felix Lucka is with the {Centrum Wiskunde \& Informatica, Amsterdam, Netherlands.}
}
\thanks{Jonas Adler is with the {Department of Mathematics, KTH - Royal Institute of Technology, Stockholm, Sweden and {Elekta, Stockholm, Sweden.}}}

\thanks{Accompanying code for this publication can be found at: \url{https://github.com/asHauptmann/3DPAT_DGD}}
}

% The paper headers
% \markboth{Journal of \LaTeX\ Class Files,~Vol.~13, No.~9, September~2014}%
% {Shell \MakeLowercase{\textit{et al.}}: Model based learning for \\ accelerated 3D photoacoustic tomography }

% make the title area
\maketitle
% As a general rule, do not put math, special symbols or citations
% in the abstract or keywords.
\begin{abstract}
Recent advances in deep learning for tomographic reconstructions have shown great potential to create accurate and high quality images with a considerable speed-up. In this work we present a deep neural network that is specifically designed to provide high resolution 3D images from restricted photoacoustic measurements. The network is designed to represent an iterative scheme and incorporates gradient information of the data fit to compensate for limited view artefacts. Due to the high complexity of the photoacoustic forward operator, we separate training and computation of the gradient information. A suitable prior for the desired image structures is learned as part of the training. The resulting network is trained and tested on a set of segmented vessels from lung CT scans and then applied to \emph{in-vivo} {photoacoustic} measurement data.
\end{abstract}

% Note that keywords are not normally used for peerreview papers.
\begin{IEEEkeywords}
Deep learning, convolutional neural networks, photoacoustic tomography, iterative reconstruction
\end{IEEEkeywords}

% For peer review papers, you can put extra information on the cover
% page as needed:
% \ifCLASSOPTIONpeerreview
% \begin{center} \bfseries EDICS Category: 3-BBND \end{center}
% \fi
%
% For peerreview papers, this IEEEtran command inserts a page break and
% creates the second title. It will be ignored for other modes.
%\IEEEpeerreviewmaketitle

\section{Introduction}
% The very first letter is a 2 line initial drop letter followed
% by the rest of the first word in caps.
% 
% form to use if the first word consists of a single letter:
% \IEEEPARstart{A}{demo} file is ....
% 
% form to use if you need the single drop letter followed by
% normal text (unknown if ever used by IEEE):
% \IEEEPARstart{A}{}demo file is ....
% 
% Some journals put the first two words in caps:
% \IEEEPARstart{T}{his demo} file is ....
% 
% Here we have the typical use of a "T" for an initial drop letter
% and "HIS" in caps to complete the first word.
%\IEEEPARstart{}{} 

Photoacoustic Tomography (PAT) is an emerging "Imaging from Coupled Physics" technique \cite{ArSc12} that can obtain high resolution 3D \emph{in-vivo} images of absorbed optical energy by sensing laser-generated ultrasound (US) \cite{Wa09,Bea11,NiXi14,Valluru2016,Zhou2016,Xia:2015sawb}. 
If data is obtained over a complete surface surrounding the domain of interest, and for all times over which the acoustic waves are propagating, then the inverse problem 
%\Marta{[remove (because it is well posed then): is only mildly illposed and]} 
can be solved directly by several analytical or numerical algorithms \cite{KuKu11}. 
%Of these methods, the fastest are of filtered-back-projection (FBP) type; see Section \ref{Sect:DirectPATrecon} for an overview of \Marta{such} methods. 
The fastest of such methods just require to solve a single wave equation; see Section \ref{Sect:DirectPATrecon} for details.
In many practical applications, restricted spatial and/or temporal sampling of the US signal is either imposed due to geometrical limitations (e.g. limited view) \cite{Xu2004}, or by the choice to utilise a compressed-sensing (CS) undersampling strategy in order to accelerate data acquisition \cite{HaNg17}. In such cases, direct reconstruction methods are not optimal to obtain high quality reconstructions as they give rise to artefacts and/or adverse noise amplification. 

Recently, several groups showed that variational image reconstruction methods that iteratively minimise a penalty function involving an explicit model of the US propagation and prior constraints on the image structure can provide significantly better results in these situations \cite{HuWaNiWaAn13,ArBeCoLuTr16,ArBeBeCoHuLuOgZh16,NaLuZhBeArBeCo17,BoLaStGiMaBr17,Schwab2018}. However, a crucial drawback of these methods is their considerably higher computational complexity and the difficulty to handcraft prior constraints that capture the spatial structure of the target accurately enough. 

As the strongest contrast in biological soft tissue is given by haemoglobin, a central promise of PAT is to deliver high quality images of blood vessel networks, e.g., for assessing the vascularization of tumors \cite{ZaVaGa14,JaLaOgTrCoZhJoPiPhMaLyPePuBe15}. Consequently we assume in this study that our targets are vessel rich and hence we learn suitable prior constraints from a set of segmented vessels.

\subsection{Deep Learning in Image Reconstruction}\label{sec:IntroDeepLearn}
The huge recent success of Deep Learning methods in image processing and computer vision has seen an increasing interest in applying similar strategies to tomographic reconstruction problems. Deep Neural Networks (DNN) are especially popular due to the low latency of a forward pass through a network which leads to prospective highly efficient reconstruction algorithms. 

In this paper we differentiate between two fundamentally different approaches to involve learning in image reconstruction:
\begin{enumerate}
\item Reconstruction followed by learning based post-processing. In this approach image reconstruction is carried out using a simple inversion step, and post-processing is used to remove artefacts and noise.
\item Model based learning and reconstruction. In this approach the forward and adjoint operators of the imaging problem are used directly in the inverse algorithm, with a multiscale regularisation scheme whose parameters are learned in the training phase.
\end{enumerate}

Many applications of Deep Learning for image reconstruction have been concentrated on the first approach by using a fast and simple %FBP-like 
direct reconstruction algorithm to obtain low quality and corrupted images and then train a convolutional neural network (CNN) on removing those artefacts, see \cite{kang2017deep,JiMcFrUn17} for an application to CT, \cite{AnHaSc17} for PAT, and MRI \cite{SaDiChVa17}. 

%{The texture-like appearance of the artefacts in images reconstructed with a simple FBP-like algorithm from sub-sampled data, has motivated training of convolutional neural networks (CNNs) for removal of such artefacts \cite{AnHaSc17,JiMcFrUn17,SaDiChVa17}. 
%While the afore-mentioned approach works , it severs the connection between the physical forward problem and the network. 

Alternatively following the second approach by including the physical forward model into the network has been studied in \cite{AdOk17,Adler2018,ChZhZhSuLiHeZhWa17,HaKlKoReSoPoKn17,kelly2017deep}. However, these improved results in reconstruction quality typically come at the cost of longer computation times which are effectively limited by the repeated simulation of the physical model.

In this paper we take the second approach. In particular, we utilize  our knowledge of the forward operator in the reconstruction process, but we will not invoke handcrafted prior constraints on the vessel structures that we are interested in. Instead, we will learn them from the data.

%The underlaying idea is to interpret a fixed small number of iterations of an iterative method for solution of the variational formulation of the problem as layers of the network. 
%An obvious possibility here is to use a FBP-like algorithm and train a DNN to remove reconstruction artefacts, this has been done with convolutional neural networks (CNN) in \cite{AnHaSc17,JiMcFrUn17,SaDiChVa17}. 

\subsection{Compressed Sensing and Limited View PAT}
In several imaging modalities the application of compressed sensing methods has been studied as a means to achieve faster acquisition speeds and/or a reduced dose when using ionising radiation \cite{Donoho:2006cs,
%Candes:2006rup,
Candes:2006ssr,Duarte:2008ricecam}. In PAT this has been studied for example in \cite{ArBeBeCoHuLuOgZh16,Provost:2009cspat,Guo:2011,Haltmeier2016, Bet17tci}. Because these methods mandate an appropriate regularisation strategy, the involvement of Deep Learning in compressed sensing is an important topic for study.

As well as data sub-sampling, in this paper we also consider the limited-view problem. Due to geometric restrictions, one can often only access the US field on one side of the tissue. A detailed examination and discussion of sub-sampling combined with the limited view problem for PAT can be found in \cite{ArBeBeCoHuLuOgZh16}.

% While the CNN based approach is promising for under-sampled full view data it is less applicable to limited view data, since the resulting artefacts are not translation invariant and a CNN is not optimal for removing them. This is discussed further below in Section \ref{sec:PAT} and demonstrated in Section \ref{sec:Experiments}. %Thus, the extension of Deep Learning to iterative methods including the physical model has attracted recent attention and showed first highly promising results \cite{AdOk17,AdOk17a,ChZhZhSuLiHeZhWa17,HaKlKoReSoPoKn17} (\Andreas{Even though none of those actually apply it to limited-view data!?}). However, these improved results in reconstruction quality typically come at the cost of longer computation times which are effectively limited by the repeated simulation of the physical model. 

%Section \ref{sec:PAT}. 

\subsection{Overview of this paper}

The rest of the paper is organised as follows. In Section \ref{sec:PAT} we discuss the physical model of photoacoustic signal generation, as well as direct reconstruction approaches, variational and the corresponding iterative reconstruction approaches, and an outline of the model based learning approach. In Section \ref{sec:Implementation} we give a detailed description of the architecture and implementation of the model based learning approach as well as a description of its training steps. In Section \ref{sec:Experiments} we discuss the measurement details, generation of training data as well as post-processing, i.e. denoising/artifact removal, of direct reconstructions. Results for simulated and 3D \emph{in-vivo} data are shown. Section \ref{sec:Discussion} provides a detailed evaluation of the results. Finally in Section \ref{sec:Conclusions} we provide some conclusions and outlook for the future.

%\Simon{Should we also here state a few definitions. For example, do we have a specific definition of a CNN or is it just assumed everyone knows what that is ? So we need to explain the difference between a DNN and  CNN, if, indeed, they are different...}

\setlength{\fboxsep}{0pt}

\begin{figure*}[t!]
\vspace{-0.5 cm}
\centering
\subfigure[][\label{subfig:TubePhantom1} volume rendering of the numerical tube phantom and the sensor locations (pink dots)]{\includegraphics[height=0.12\textwidth]{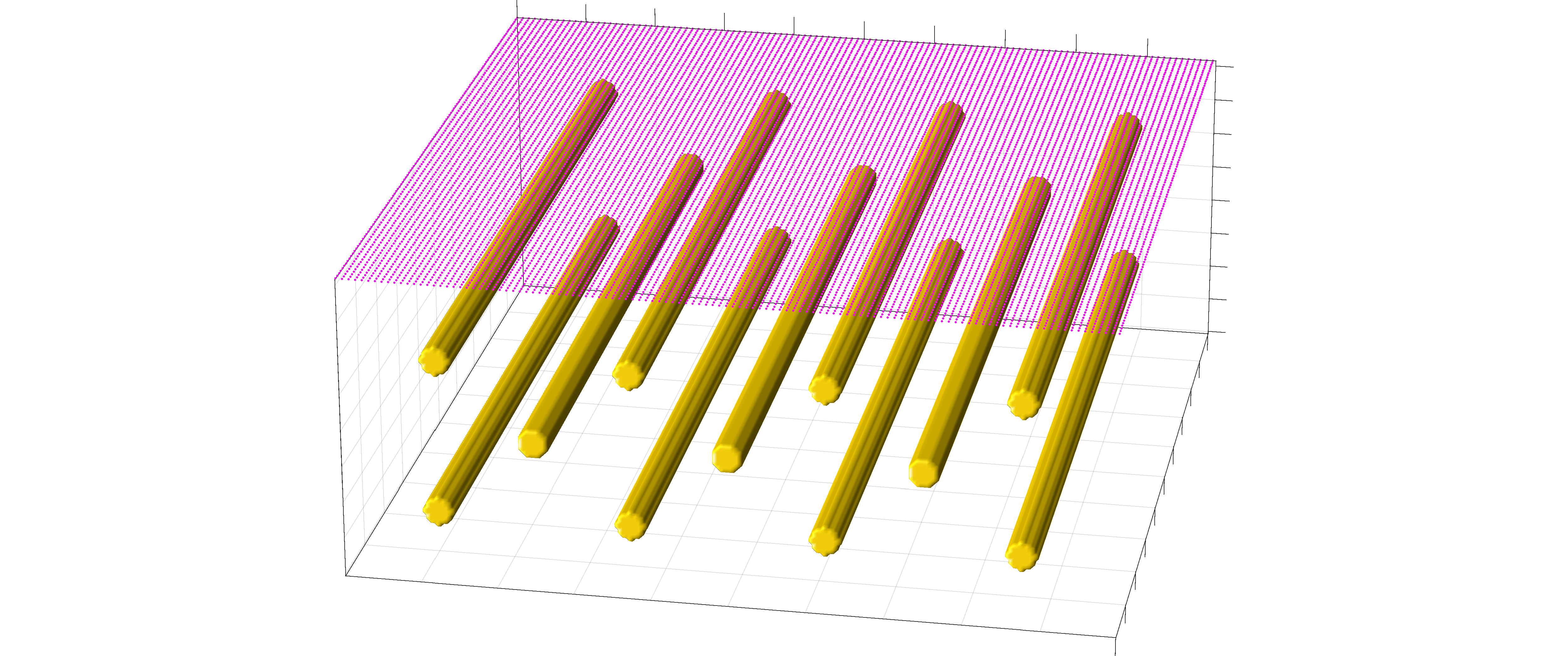}}
\hspace{-5.35 cm}
\hfill
\subfigure[][\label{subfig:TubePhantom2}slice view through the tube phantom]{\includegraphics[height=0.12\textwidth]{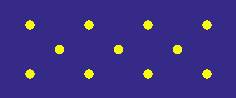}}
\hfill
{\vspace{-3.35 cm} }\\
\subfigure[][$A^* y$, full data \label{subfig:TubesBPfull}]{\includegraphics[height=0.12\textwidth]{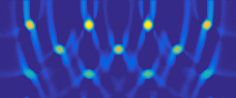}}
\hfill
\subfigure[][$A^* y$, sub-sampled data \label{subfig:TubesBPSS}]{\includegraphics[height=0.12\textwidth]{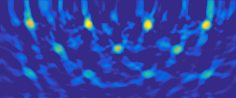}}
\hfill
\subfigure[][sub-sampling pattern \label{subfig:SSPat}]{\fbox{\includegraphics[height=0.29\textwidth]{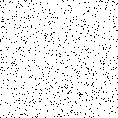}}}\\
\subfigure[][NNLS, 5 iter, full data \label{subfig:TubesNNLS5full}]{\includegraphics[height=0.12\textwidth]{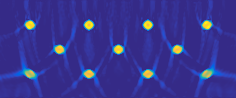}}
\hfill
\subfigure[][NNLS, 5 iter, sub-sampled data\label{subfig:TubesNNLS5SS}]{\includegraphics[height=0.12\textwidth]{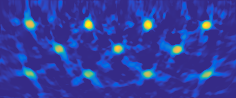}}
\hfill
\subfigure[][TV, 5 iter, sub-sampled data\label{subfig:TubesTV5SS}]{\includegraphics[height=0.12\textwidth]{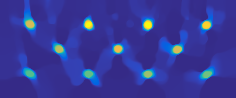}}\\
\subfigure[][NNLS, 20 iter, full data\label{subfig:TubesNNLS20full}]{\includegraphics[height=0.12\textwidth]{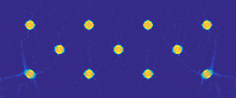}}
\hfill
\subfigure[][NNLS, 20 iter, sub-sampled data\label{subfig:TubesNNLS20SS}]{\includegraphics[height=0.12\textwidth]{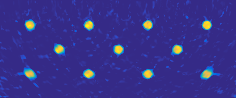}}
\hfill
\subfigure[][TV, 20 iter, sub-sampled data\label{subfig:TubesTV20SS}]{\includegraphics[height=0.12\textwidth]{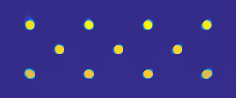}} \\
{\includegraphics[width=\textwidth]{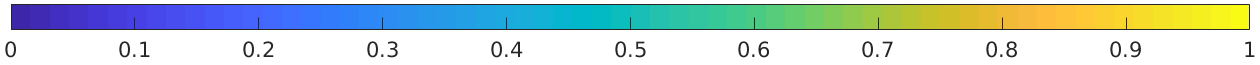}}
\vspace{-0.5 cm}
\caption{\label{fig:Tubes} Illustration of the properties and errors of different image reconstruction methods using a simple numerical phantom consisting of tubes. \protect\subref{subfig:TubePhantom1}\&\protect\subref{subfig:TubePhantom2}: Visualizations of the numerical phantoms. \protect\subref{subfig:SSPat}: Illustration of the sub-sampling pattern. Every pixel corresponds to one of the $118\times118$ scanning locations shown as pink dots in \protect\subref{subfig:TubePhantom1}. We sub-sample by a factor of $16$, i.e., of all locations, a fraction of $1/16$ is chosen at random and visualized by a black pixel. \protect\subref{subfig:TubesBPfull}-\protect\subref{subfig:TubesBPSS}\&\protect\subref{subfig:TubesNNLS5full}-\protect\subref{subfig:TubesNNLS20SS}: Slice views through the reconstructions of the tube phantom by different methods and for full or sub-sampled data.}
\end{figure*}

%--------------------------------------------------------------------
\section{Photoacoustic Tomography}\label{sec:PAT}
%--------------------------------------------------------------------

\subsection{Photoacoustic Signal Generation}
To generate the PA signal, a short pulse of near-infrared laser light is sent into biological tissue where the photons will get scattered and absorbed by any chromophores present. Under certain conditions (see \cite{WaAn11} for details), part of the absorbed optical energy will be \textit{thermalised}, i.e., converted to heat, and the induced local pressure increase $x$ 
%(usually referred to as PA image) 
travels through the tissue as an US wave (\textit{photoacoustic effect}). Spatio-temporal measurements of these waves at the boundary of the tissue constitute the PA signal $y$. A common way to model the acoustic part of the signal generation is to consider the following initial value problem for the wave equation \cite{KuKu11,ArBeCoLuTr16,WaAn11},
\begin{equation}
(\partial_{tt} - c_0^2 \Delta) p(r,t) = 0, \quad p(r,t = 0) = x, \quad \partial_t p(r,t = 0) = 0.\label{eqn:PATfwd}
\end{equation}
The US sensing is then modeled as a linear operator $\mathcal{M}$ acting on the pressure field $p(r,t)$ restricted to the boundary of the computational domain $\Omega$ and a finite time window (see \cite{Bea11,LuRa13} for details on measurement systems):
\begin{equation}
y = \mathcal{M} \, p_{|\partial \Omega \times (0,T)}. \label{eqn:Measurement}
\end{equation}
Equations (\ref{eqn:PATfwd}) and (\ref{eqn:Measurement}) define a linear mapping 
\begin{equation}
Ax=y,
\label{eqn:Axeqy}
\end{equation}
from initial pressure $x$ to measured pressure time series $y$, which constitutes the \emph{forward problem} in PAT. The corresponding image reconstruction step constitutes the \emph{inverse problem} to (\ref{eqn:Axeqy}). 
%The discretization of this mapping casts this first, purely acoustic part of PAT
%\Marta{[do you mean that we could do QPAT but do not? I would drop any hints to
%what we are not doing]} into the form of eq.(\ref{eqn:Axeqy}). Obtaining a high
%quality estimate of $x$ from $y$ is crucial for any kind of PAT applications but %in particular for quantitative PAT (QPAT) \cite{CoLaArBe12}, where the aim is to %also invert the (non-linear) optical part of the signal generation based on the %PA image. \\

%\Simon{
Note that $x$ is a $N_x \times N_y \times N_z$ 3D image of initial pressure and $y$ is a $N_h \times N_v \times N_t$ volume of acquired pressure data as a function of acoustic propagation time. 
In the examples used in this paper this results in dimension of $A$ of around $7M$ by $4.6M$ which (if fully dense) would require about 123TB of memory in single precision which is intractable for currently available computational resources. Thus image reconstruction methods require either direct, or iterative "matrix-free" implementations as discussed in the next sections.
%}

\subsection{Direct methods for PAT Image Reconstruction}\label{Sect:DirectPATrecon}

%\felix{[In the original version of this section, I did not mention time reversal at all to keep things simple and because we don't need it later or show any time reversal results. I would recommend to drop any references to it, because the reviewers might otherwise ask for TR reconstructions or whether using TR + U-Net is better...it distracts from the main story.]}

Direct methods are especially attractive in the large scale setting as they only require solution of a single wave equation; i.e., given a computational solver for \eqref{eqn:PATfwd} we can compute an inverse solution with the same computational cost \cite{ArBeCoLuTr16}. In particular in this study we choose to compute the \emph{adjoint solution} $A^* y$, which is close to the inverse solution.

%or the \emph{time reversed solution} $A^{\triangleleft} y$ \cite{FiPa04,XuWa04b,TrZhCo10}}. Both inverse solvers have the same computational cost as the forward solver; for a discussion of the difference between adjoint and time-reversal approaches we refer to \cite{ArBeCoLuTr16}.

Here, as the wave solver we use a pseudo-spectral time-domain method 
\cite{MaSoLiTaNaWa01,CoKaArBe07,TrZhCo10} as implemented in the \kwave/ Matlab Toolbox \cite{TrCo10}, which allows to run the computations on GPU cards using fast CUDA code. 
%The details of this implementation can be found in \cite{ArBeCoLuTr16}. 
%With these preparations, one can employ different methods to solve \eqref{eqn:Axeqy}: Linear, back-projection-type approaches solve a single wave equation backwards in time. For instance, applying $A^*$ to $y$ can be expressed in this way \cite{ArBeCoLuTr16}, but more prominently, \textit{time reversal} approaches \cite{FiPa04,XuWa04b,TrZhCo10} rely on this formulation. \Simon{Need to make these statements consistent with the "FBP-like" comments in the introduction}. 

Whilst direct approaches are computationally efficient they are inadequate for dealing with the sub-sampled limited-view data employed in this paper as we demonstrate next. %illustrate next.
Figure \ref{fig:Tubes} illustrates the influences of limited-view and sub-sampling on a simple numerical phantom of tubes that should mimic blood vessels. From Figure \ref{subfig:TubesBPfull}, we can see that a reconstruction by $A^* y$ suffers from severe circular artefacts \cite{FrQu15} and a systematic loss of contrast with depth. Figure \ref{subfig:TubesBPSS} shows that these problems {are accentuated} with sub-sampled data. 
%\Simon{Similar artefacts appear if using time-reversal, with somewhat worse accuracy.}

\subsection{Variational approach to PAT image reconstruction}\label{sec:PATvariational}

Variational methods aim to recover the PA image $x$ in \eqref{eqn:Axeqy} from the measurement $y$ as a minimiser of a penalty function, 
\begin{equation}\label{eqn:penaltyFunctional}
x \in \argminsub{x'} \left\lbrace J(x') \right\rbrace = \argminsub{x'} \left\lbrace d(y,Ax')+\lambda R(x') \right\rbrace,
\end{equation}
where the fidelity term $d(y,Ax')$ measures the data fit and a regularising term $R(x)$ encodes prior knowledge about the structures in the image. Often, $R(x)$ is convex but not differentiable. A simple approach to find a solution to \eqref{eqn:penaltyFunctional} is given by a \textit{proximal-gradient-descent} scheme:
\begin{equation}
x_{k+1} = \text{prox}_{R,(\lambda \gamma_{k+1})} \left(x_k - \gamma_{k+1} \nabla d(y,Ax_k) \right), \label{eqn:ProxGrad}
\end{equation}
\revision{with step length $\gamma>0$ and} where the \textit{proximal operator} solves an image denoising problem:
\begin{eqnarray}
\text{prox}_{R,\alpha}(y) = \argminsub{x} \left\lbrace R(x) + \frac{1}{2 \alpha} \sqnorm{x - y} \right\rbrace. \label{eqn:ProxOp}
\end{eqnarray}
The drawback of the above procedure is the difficulty to choose a suitable regularisation term $R(x)$, a regularisation parameter $\lambda>0$, that balances data fit and the regularisation properties, and the potentially large number of iterations it takes to converge.

As shown in, e.g., \cite{HuWaNiWaAn13,ArBeCoLuTr16,ArBeBeCoHuLuOgZh16}, iterative image reconstruction methods of the form \eqref{eqn:ProxGrad} that solve variational regularisation problems \cite{ScGrGrHaLe09} like \eqref{eqn:penaltyFunctional} can improve upon {the direct image reconstruction methods}. For instance, we can incorporate the physical constraint that the initial pressure increase $x$ is always positive by choosing $R(x)$ to be $0$ if $x \geqslant 0$ and $\infty$ otherwise. For this, $\text{prox}_{R,\alpha}(y) = \max(y,0)$. With the canonical choice $d(y,Ax') = \tfrac{1}{2} \|Ax'-y\|_{2}^2$, \eqref{eqn:penaltyFunctional} simply becomes a non negative least squares (NNLS) solution. Figures \ref{subfig:TubesNNLS5full}, \ref{subfig:TubesNNLS20full} demonstrate that with increasing number of iterations, both limited-view artefacts and the systematic loss of contrast disappear. However, they also show that the convergence in deeper, non-central parts of the image is considerably slower and the limited-view will still manifest in blurry edges. For the sub-sampled data case shown in Figures \ref{subfig:TubesNNLS5SS}, \ref{subfig:TubesNNLS20SS} we see similar effects although in addition, noise-like artefacts remain. As examined in \cite{ArBeBeCoHuLuOgZh16}, using noise-reducing, edge-preserving regularisation like the \revision{(isotropic)} total variation (TV) functional $R(x) = \|\nabla x\|_1$ can further improve such results as can be seen in Figures \ref{subfig:TubesTV5SS}, \ref{subfig:TubesTV20SS}. The main problem of such iterative approaches is in terms of computation times, compared to the linear backprojection by $A^* y$ which requires the solution of one wave equation, computing $20$ iterations of NNLS or TV requires in total $40$ additional solutions of a wave equation.

\setlength{\fboxsep}{0pt}

\begin{figure*}[t!]
\centering
\begin{picture}(420,125)
\put(10,0){\includegraphics[width=400 pt]{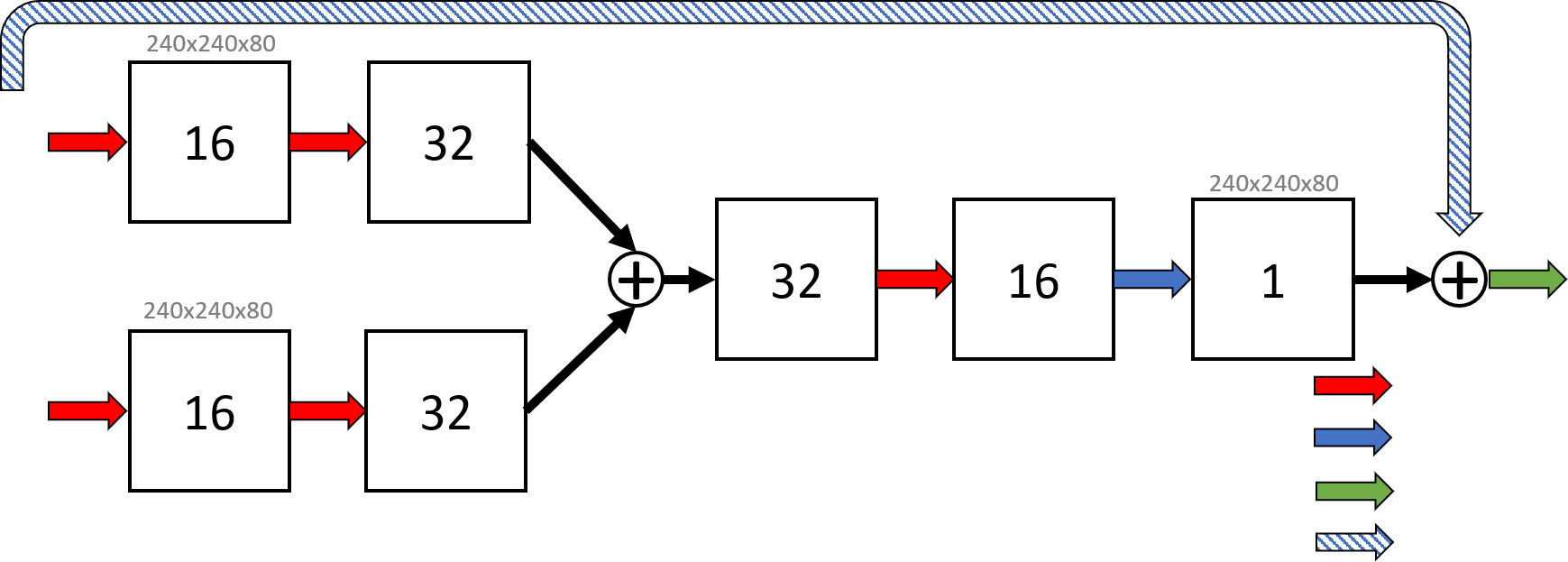}}
\put(10,106){$x_k$}
\put(-27,37){{$\nabla d(y,Ax_k)$}}
\put(370,42.5){\small ReLU(conv$_{5\times5\times5}$)}
\put(370,29){\small $\lambda\cdot$(conv$_{5\times5\times5}$)}
\put(370,15){\small ReLU}
\put(370,2.5){\small skip connection}
%\put(162.5,92){\huge \revision{Adjust, how?}}
%\put(175.8,51.1){\huge $+$}
%\put(377.4,51.1){\huge $+$}
%\put(382,18){$x_k$}
\put(415,70){$x_{k+1}$}
\end{picture}
\caption{\label{fig:gradientFlowGraph} Diagram of one convolutional neural network, denoted as $G_{\theta_k}$, representing one iteration of the deep gradient descent. \revision{Image size for input and output is indicated in gray.} The red arrows denote a convolutional layer with $5\times5\times5$ kernels followed by a $\mathrm{ReLU}$, \revision{the resulting channels in each layer are indicated in the squares}. The blue arrow denotes a convolutional layer followed by a scalar multiplication. \revision{The residual update (by the skip connection) is then projected to the positive numbers by the last $\mathrm{ReLU}$.}}
\end{figure*}
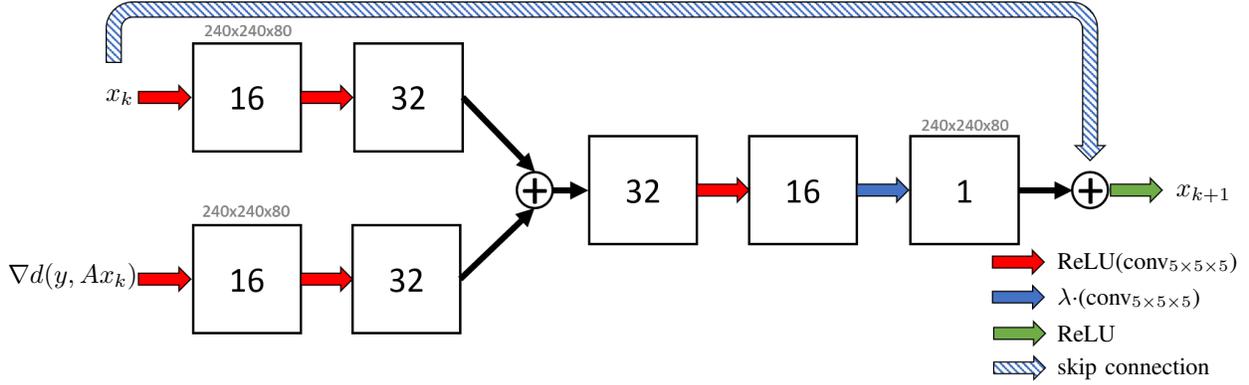

\subsection{Model Based Learning}
Regularisation functionals like TV are popular because they often allow for a mathematical analysis of the minimisers of \eqref{eqn:penaltyFunctional} and have been designed to perfectly recover certain aspects of $x$, e.g., its singularities \cite{BuOs13}. As such, they often yield spectacular results for simple numerical or experimental phantoms like the ones shown in Figure \ref{fig:Tubes}. In many applications however, typical images $x$ have a more involved structure and the prior information expressed by simple regularisers like TV does not lead to optimal results. One example is given by sub-sampled PAT measurements of vessel networks \cite{ArBeBeCoHuLuOgZh16}. If we have a set of typical PA images of vessel networks, we could try to learn more suitable prior information and how to best incorporate it in an iterative image reconstruction approach that also utilizes measurement information over the gradient of the data fit,
\begin{equation}
\nabla \tfrac{1}{2} \|A x_k - y\|_{2}^2 = A^*(A x_k - y),
\end{equation}
at every step $k$. 

Inspired by \cite{AnDeGoHoPfScFr16,PuWe17} we take the structure of \eqref{eqn:ProxGrad} as a starting point: Each iteration consists of updating $x_k$ by combining measurement information delivered through the gradient $\nabla d(y,Ax_k)$ with an image processing step. Instead of deriving the concrete form of this combination from a fixed penalty function \eqref{eqn:penaltyFunctional}, we propose to learn instead an update function for each iteration
\begin{equation}\label{eqn:learnedUpdate}
x_{k+1}={G}_{\theta_k}(\nabla d(y,Ax_k),x_k).
\end{equation}
This implies that the effect of the regularising term is now learned from the data during training. The functions $G_{\theta_k}$ correspond to CNNs with different, learned parameters $\theta_k$ but with the same architecture. The network structure is kept simple and should mimic a proximal gradient update \eqref{eqn:ProxGrad}. Due to the representation of each update by a CNN applied to the current $x_k$ and the gradient $\nabla d(y,Ax_k)$, we call the whole algorithm \emph{deep gradient descent} (DGD).

In contrast to \cite{AdOk17,HaKlKoReSoPoKn17,PuWe17} {we train the DGD layer by layer (layer corresponding here to one iterate),  i.e.~we learn the parameters $\theta_k$ for each iteration separately. In this way we can exclude the photoacoustic operator from the training procedure. This is necessary to make the training feasible. Note, that the photoacoustic operator has complexity $\mathcal O(N^4 \log(N))$
\cite{ArBeCoLuTr16}, for a volume of size $n = N \times N \times N$, compared to CT and MRI where $A$ has complexity $\mathcal O(N^3 \log(N))$ for a volume of size $n = N \times N \times N$. Therefore we think that such layer by layer training scheme is the only feasible approach for iterative high-resolution 3D PAT imaging at the present stage.}

%--------------------------------------------------------------------
\section{Implementation\label{sec:Implementation}}
%--------------------------------------------------------------------
\revision{In a CNN, each layer is of the following form: Given the input $g$ and output $h$ with channel index sets $\mathcal{I}$, $\mathcal{J}$ respectively, then 
\[
h_i=\varphi\left(b_i + \sum_{j\in \mathcal{J}} \omega_{i,j} \ast g_j \right), \ i \in \mathcal{I},
\]
with a componentwise nonlinear function $\varphi$ and convolution $\ast$. The whole parameter set $\theta$ of the network is therefore given by the biases $b_i\in\R$ and convolutional filters $\omega_{i,j}\in\R^{s^n}$ (with kernel size $s$ and spatial dimension $n$) of each network layer.}

The specific architecture we have chosen for the CNNs performing the update in equation \eqref{eqn:learnedUpdate} is illustrated in Figure \ref{fig:gradientFlowGraph}. In each iteration we input $x_k$ and $\nabla d(y,Ax_k)$ to a similar pipeline, where both are spread to 16 and then 32 channels by a convolutional layer with \revision{kernel size $s=5$ and dimension $n=3$, equipped with a rectified linear unit as nonlinearity,} that is defined as 
\[
\mathrm{ReLU}(x)=\max(x,0).
\]
The output of both pipelines is added together and first reduced to 16 \revision{channels, equipped with a $\mathrm{ReLU}$, and then to 1 channel without a nonlinearity, but a simple scalar multiplication.} The result is added to the current iterate and projected to the positive numbers by a $\mathrm{ReLU}$, similar to the proximal for NNLS discussed in Section \ref{sec:PATvariational}.

\revisNew{The architecture in this study is motivated by a typical network structure consisting of an analysis/encoding and a reassembling/decoding part. In this analysis part, the number of channels is increased between layers to refine the analysis of the features extracted in the layer before. In the reassembling part, these features must be merged/thresholded to produce an output image, so the number of channels is decreased.} 
\revisNew{Since the main contribution of this work is not the specific neural network architecture, we use a simple architecture following this convention.} \revision{In particular, the network structure is kept rather small with the motivation in mind that each $G_{\theta_k}$ primarily learns how to combine current iterate and gradient as well as a data specific filters, in contrast to a large post-processing network. Furthermore, a compact structure is necessary to minimise the needed memory on the GPU.}

\subsection{Training of the deep gradient descent}
Given a training set $\{y^i,x_{true}^i\}_i$, we have two options to train the parameters $\theta_k$. The first is to pre-define a maximum of iterations $k_{max}$ and train all $\theta_k$, for $k=0,\dots,k_{max}-1$,\linebreak together to minimise the difference between $x_{true}^i$ and the result of the last iteration $x_{k_{max}}^i$; that is we seek to find
\begin{equation}\label{eqn:TrainTargetFull}
\mathcal E_{k_{max}} = \min_{\theta_{0},\dots, \theta_{k_{max}-1}} \sum_i \| x_{k_{max}}^i - x_{true}^i \|,
\end{equation}
for some suitable norm.
The second approach is to train the parameters sequentially: $\theta_0$ is trained to minimise the difference between $x_{true}^i$ and $x_1^i$ given data $y^i$, for all indices $i$. After that, $\theta_1$ is trained to minimise the difference between $x_{true}^i$ and $x_2^i$, given the optimal $x_1^i$ found in the training of the first CNN $G_{\theta_0}$. That means the minimisation of \eqref{eqn:TrainTargetFull} is split into $k_{max}$ independent optimisation problems w.r.t.~disjoint subsets of parameters $\theta_k,\ k=0,\dots, k_{max}-1$, given by
\begin{equation}\label{eqn:TrainTargetSeq}
\min_{\theta_k} \sum_i \| x_{k+1}^i - x_{true}^i \|, \quad x_{k+1}^i = G_{\theta_k}(\nabla d(y,Ax_{k}^i),x_{k}^i).
\end{equation} 

The first approach has the advantage that the network is more flexible to achieve the best possible result after $k_{max}$ iterations, but during the training, the operators $A$ and $A^*$ need to be evaluated many times, \revision{since for each training step all $x_k$ and their corresponding gradients have to be computed to evaluate \eqref{eqn:TrainTargetFull}.}
While the second approach is not optimal in the sense that it does not lead to minimal training error, it has two important advantages. Firstly, the computation of the gradient $A^*(A x - y)$ and training decouple, which is important in view of the cost of application of $A$ and $A^*$ in PAT. Secondly, it provides an upper bound on the training error \eqref{eqn:TrainTargetFull}.
In fact, \eqref{eqn:TrainTargetSeq} can be viewed as a greedy approach which seeks to obtain a minimum in each layer $k$ given $x_{k-1}$ from the previous training step.
We note that this property can be used to determine the number of layers $k_{max}$ of the DGD in training by controlling the training error from layer to layer in contrast to choosing it a priori. Therefore, the second approach could also be used as a pre-training stage to initialize the weights for the first approach.

As the computational complexity of simulating acoustic wave propagation in 3D prohibits computing the gradient during any training scheme, we need to follow the second approach here. The whole training procedure we use is summarized in Algorithm \ref{alg:algorithmTrain} for a given number of maximum iterations $k_{max}$ and the reference solution $x_{true}$.

\begin{algorithm}
	\caption{Training Procedure}
	\label{alg:algorithmTrain}
	\begin{algorithmic}[1]
		\Let{$x_0$}{$A^*y$}
        \Function{trainingCycle}{}
        \Let{$k$}{$0$}
        \While{$k<k_{max}$} 
        \State Compute $\nabla d(y,Ax_k)=A^*(Ax-y)$
        \Function{trainIterate}{$\nabla d(y,Ax_k),x_k,x_{true}$}
        \State Train for given accuracy
        \EndFunction{(\Return{${\theta_k}$})}
        \Let{$x_{k+1}$}{$G_{\theta_k}(\nabla d(y,Ax_k),x_k)$}
        \Let{$k$}{$k+1$}        
        \EndWhile
        \EndFunction
	\end{algorithmic}
\end{algorithm}

\subsection{Evaluation of the deep gradient descent}
After training the parameter sets $\{\theta_k\}_{k=0}^{k_{max}-1}$, the learned iterative reconstruction scheme can be evaluated as follows: The new iterate $x_{k+1}$ is computed by applying the network $G_{\theta_k}$ to the current iterate $x_k$ and the gradient of the data fit, in particular this means that the gradient has to be computed in every iteration. This procedure is equivalent to Algorithm \ref{alg:algorithmTrain} without calling \revision{$\textsc{trainIterate}$} in line 6-8. 

% \begin{algorithm}
% 	\caption{Evaluation Procedure}
% 	\label{alg:algorithmEval}
% 	\begin{algorithmic}
% 		\Let{$x_0$}{$A^*y$}
%         \Function{evaluationCycle}{}
%         \For{$k=0:k_{max}$} 
%         \State Compute $\nabla d(x_k)=A^*(Ax-y)$
%         \Let{$x_{k+1}$}{$G_{\theta_k}(\nabla d(x_k),x_k)$}
%         \EndFor
%         \EndFunction
% 	\end{algorithmic}
% \end{algorithm}

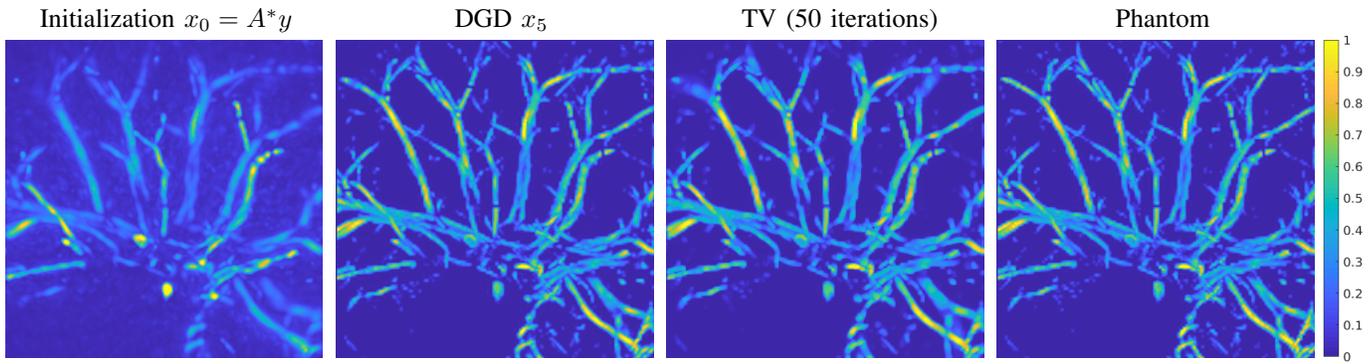
\begin{figure*}[t!]
\centering
\begin{picture}(500,140)
\put(00,0){\includegraphics[width=120 pt]{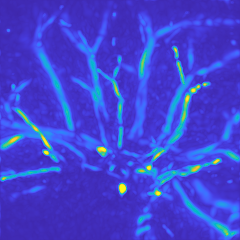}}
\put(125,0){\includegraphics[width=120 pt]{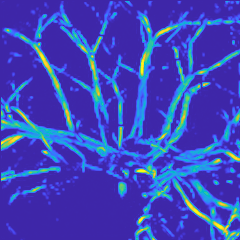}}
\put(250,0){\includegraphics[width=120 pt]{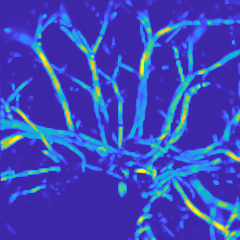}}
\put(375,0){\includegraphics[width=120 pt]{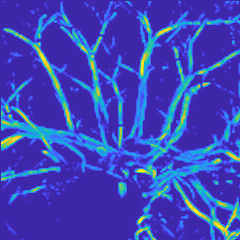}}
\put(497,-2){{\includegraphics[height=125 pt]{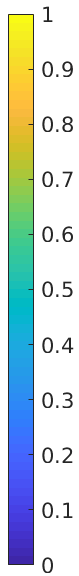}}}
\put(13,125){Initialization $x_0=A^*y$}
\put(170,125){DGD $x_5$}
\put(275,125){\revisNew{TV (50 iterations)}}
\put(420,125){Phantom}

\end{picture}
\caption{\label{fig:SimulationResult} Reconstruction results for a test image from the segmented CT data (not included in the training), presented images are top-down maximum intensity projections. \revisNew{From left to right: Backprojection of the data and initialization of the network, result of DGD with 5 iterations, TV reconstruction with 50 iterations, phantom used to produce the data.}}
\end{figure*}

%--------------------------------------------------------------------
\section{Experiments}\label{sec:Experiments}
%--------------------------------------------------------------------
In this study we are interested in reconstructing human {\it in-vivo} data and hence we do not have a true target available for the training of measured data. This lack of a ground truth is one of the main challenges in supervised learning. Nevertheless, we chose to train the DGD with supervised learning using simulated data and hence a meaningful data set is crucial for a successful training, for that purpose we use %have 
segmented human vessel structures from CT scans as discussed in the next section.
%\felix{[I commented out the part about kWave, we said that in Sec 2 B already]}
%All computations that need the simulation of the photoacoustic operator have been implemented with \kwave/ in MATLAB, see \cite{TrCo10} for details. 
The training and evaluation of each network $G_{\theta_k}$ has been implemented with TensorFlow \cite{tensorflow2015-whitepaper} in Python. All computations are done on a Titan Xp GPU with 12GB memory.

\subsection{Training on segmented lung vessels}\label{sec:trainOnVessel_clean}
The training data needs to be as realistic as possible to provide a meaningful basis for the algorithm. To achieve this we have used the publicly available data from \emph{ELCAP Public Lung Image Database\footnote{http://www.via.cornell.edu/databases/lungdb.html}}. The data set consists of 50 whole-lung CT scans, from which we have segmented about 1200 volumes of vessel structures with a Frangi vesselness filter \cite{MaNi05,FrNiViVi98}. The segmented volumes were of size $40\times 120 \times 120$, and were then scaled up by a factor of 2 to the final target size of $80\times 240 \times 240$. Out of these volumes we chose 1024 as ground truth $x_{true}$ for the training and simulated limited-view, sub-sampled data using the same measurement setup used in the in-vivo data: We assume that each voxel has the isotropic length $dx = 84.75\mu$m and that the full data is recorded at locations on a grid with grid size $2dx$ on one of the two $240\times240$ sized outer planes of the volume (i.e., the scanning geometry is similar to Figure \ref{subfig:TubePhantom1}). In time, $N_t = 486$ pressure samples are recorded with $dt = 16.6$ns. The full data is then sub-sampled as illustrated in Figure \ref{subfig:SSPat} but by a sub-sampling factor of 4. We have added normally distributed noise to the measured data, such that the resulting SNR was %about 
{approximately} 15 for all measurements and we assumed a sound speed of $c_0 = 1580$m/s.
In a nutshell, we obtained the data $y=A x_{true}+\varepsilon$, where $\varepsilon$ denotes the added noise. 

The training set for one CNN \revision{$G_{\theta_k}$} then consists of current iterate $x_k$, the gradient of the data fit $\nabla d(y,Ax_k)=A^*(Ax_k-y)$, and the ground truth $x_{true}$. 
%For the initialization of the first iterate we set
{We initialize the iteration with}
\[ 
x_0=A^*y.
\]
Precomputing the gradient information for each CNN takes about 10 hours. 

We trained the CNNs using TensorFlow's implementation of Adam \cite{kingma2014adam}. 
%\Marta{Training of each network layer takes approximately 10 hours using [stochastic descent/Adam? what method?]. 
For the training we used batches of size 2,  since this already fills up the memory (12GB) of the GPU completely. We trained each $G_{\theta_k}$ for 25000 iterations {(i.e.~approximately 50 epochs)} with an initial step size of $5\cdot 10^{-5}$ (learning rate), %that is about 50 epochs. 
The minimised loss function, \revision{i.e. the norm in \eqref{eqn:TrainTargetSeq}}, is chosen as the $\ell^2$-distance of new iterate to the true solution $x_{\mathrm{true}}$, 
\[
\mathrm{loss}(x)=\|x-x_{{true}}\|_2^2.
\]
For training the first CNN $G_{\theta_0}$ we added an additional constraint to avoid the local minima of zero solutions by penalizing a small norm
\[
\mathrm{loss}_{add}(x)=-\alpha \min (\|x\|_2-\beta,0),
\]
\revision{with small $\alpha, \beta>0$.} The training of each CNN $G_{\theta_k}$ took about 1 day on the GPU. We have trained 5 iterates, i.e. $k_{max}=5$, for the deep gradient descent. In total the whole training took 7 days. \revision{We note, that this could be speed up by initialisation of $\theta_k$ with $\theta_{k-1}$ or by more advanced optimisation strategies, see for instance the review \cite{vidal2017mathematics}.}
At this point we would like to note, that %if we would have 
{had we} included the operator $A$ and $A^*$ in the training and trained all 5 iterates together, then the time needed for 25000 iterations would 
be in the order of 70 days, \revision{and used at least 5 times more memory}. The result of the DGD for simulated data is shown in Figure \ref{fig:SimulationResult} for an example that was not included in the training set.

\begin{figure}[t!]
\centering
\begin{picture}(260,130)
\put(00,0){\includegraphics[width=125 pt]{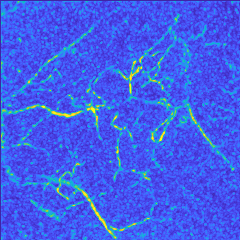}}
\put(130,0){\includegraphics[width=125 pt]{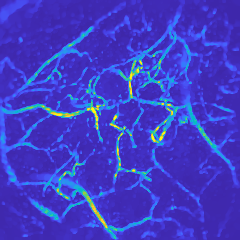}}
\put(45,130){DGD $x_5$}
\put(135,130){TV (full data), $\lambda=2\cdot 10^{-4}$ }
\end{picture}
\caption{\label{fig:realData_cleanNet} Reconstruction from real measurement data of a human palm, without adjustments of the training data. The images shown are top-down maximum intensity projections. Left: Result of the DGD trained on images without added background. Right: TV reconstruction as reference from fully sampled data.}
\end{figure}

\begin{figure*}[t!]
\centering
\begin{picture}(500,310)
\put(0,170){\includegraphics[width=140 pt]{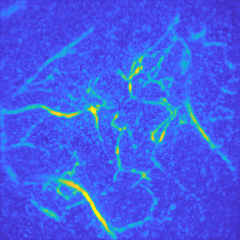}}
\put(160,170){\includegraphics[width=140 pt]{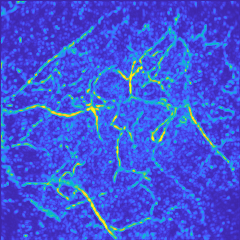}}
\put(320,170){\includegraphics[width=140 pt]{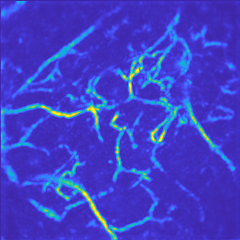}}
\put(0,0){\includegraphics[width=140 pt]{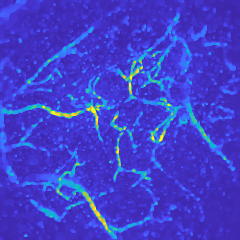}}
\put(160,0){\includegraphics[width=140 pt]{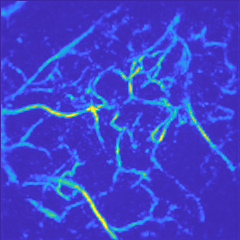}}
\put(320,0){\includegraphics[width=140 pt]{imagesRealData/RealData_Ex1_TVFull_mIX.png}}
\put(15,320){Initialization $x_0=A^*y_{real}$}
\put(215,320){DGD ${x}_5$}
\put(355,320){Updated DGD $\hat{x}_5$}
\put(10,155){TV sub-sampled, $\lambda=5 \cdot 10^{-5}$}
\put(45,143){20 Iterations}
\put(200,150){\revisNew{Updated U-Net}}
%\put(185,155){TV sub-sampled, $\lambda= 10^{-4}$}
%\put(215,143){20 Iterations}
\put(316,155){TV fully-sampled data, $\lambda=2\cdot 10^{-4}$ }
\put(365,143){20 Iterations}
\put(465,-7){{\includegraphics[height=324 pt]{images/ParulaHorzNumbers.png}}}
\end{picture}
\caption{\label{fig:RealDataResult} Example for real measurement data of a human palm. The images shown are top-down maximum intensity projections. First row: from left to right, the initialization from sub-sampled data, the output of DGD trained on background added data after 5 iterations, and updated DGD ${G}_{\hat{\theta}_k}$ after 5 iterations.  
Second row: from left to right, TV reconstruction of sub-sampled data with a emphasis on the data fit, \revisNew{updated U-Net reconstruction,} reference TV reconstruction of fully-sampled limited-view data. All TV reconstructions have been computed with 20 iterations.}
\end{figure*}

\subsection{Post-processing by Deep Learning}
To complement this study, we have also implemented the first approach of learning in image reconstruction, see Section \ref{sec:IntroDeepLearn}, \textit{viz.} taking an initial direct reconstruction and train a network to remove artefacts and noise. 
Especially popular for improving these initial reconstructions is a CNN introduced as U-Net \cite{RoFiBr15}. We refer to the original paper %on 
{for} the architecture, but roughly summarized its strength lies in a series of skip connections in a multilevel decomposition. For our application, we have followed the modified U-Net architecture proposed by \cite{JiMcFrUn17} for \revision{post-processing of} 2D X-ray tomography, that learns to compute an update to the initial reconstruction. We made the necessary modifications for a three-dimensional setting and implemented training and evaluation with TensorFlow. 

To be consistent with the previous section our direct reconstruction, which we seek to improve upon, is obtained by the application of the adjoint
$x_0=A^*y$. %as corrupted initial reconstruction. 
The modified U-Net is then trained on the %training data 
{set of pairs} $\{x_0^i,x_{true}^i\}_i$. Due to memory restrictions we were only able to 
train one pair at a time. The loss function is chosen as the combination $\mathrm{loss}(x)+\mathrm{loss}_{add}(x)$, see previous section. The training is then performed with Adam for 75 epochs and a learning rate of $10^{-4}$; this took 3 days. 
The results for simulated data will be discussed in Section \ref{sec:compPostProc}.

\begin{figure*}[t!]
\centering
\begin{picture}(500,310)
\put(0,170){\includegraphics[width=140 pt]{imagesSimData/SimData_Ex1_input_mIX.png}}
\put(170,170){\includegraphics[width=140 pt]{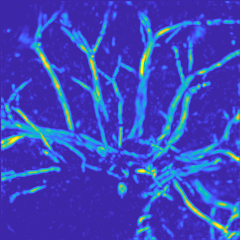}}
\put(340,170){\includegraphics[width=140 pt]{imagesSimData/SimData_Ex1_iter5_mIX.png}}
\put(25,320){Initialization $x_0=A^*y$}
\put(220,320){Iterate $x_1$}
\put(390,320){Iterate $x_5$}

\put(0,0){\fbox{\includegraphics[width=140 pt]{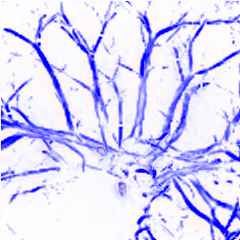}}}
\put(170,0){\fbox{\includegraphics[width=140 pt]{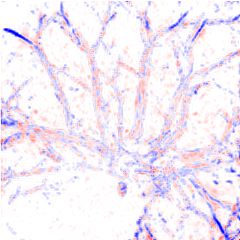}}}
\put(340,0){\fbox{\includegraphics[width=140 pt]{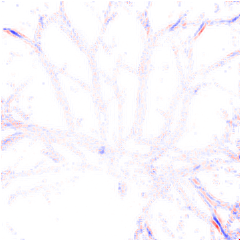}}}

\put(25,150){Difference: $x_0-x_{true}$}
\put(200,150){Difference: $x_1-x_{true}$}
\put(365,150){Difference: $x_5-x_{true}$}

\put(485,-3){{\includegraphics[height=146 pt]{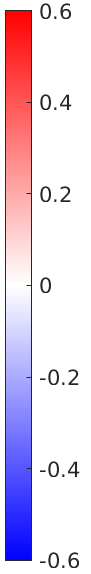}}}
\put(485,167){{\includegraphics[height=146 pt]{images/ParulaHorzNumbers.png}}}
\end{picture}
\caption{\label{fig:ProgressIterations} Progress of iterations in the DGD for a test image from the segmented CT data. 
Images shown are top-down maximum intensity projections. The top row shows reconstructions and the bottom row shows difference images to the true solution. Difference images are on the same scale, with blue for a negative difference and red for positive. Left: initialization and input to the DGD, maximal value of difference is $0.8492$. Middle: output after one iteration with DGD, maximal value of difference is $0.6171$. Right: result after 5 iterations of DGD, maximal value of difference is $0.4124$.}
\end{figure*}

\subsection{Application to in-vivo data}\label{sec:TrainRealData}

We now apply our method to in-vivo data of a human palm. The details of the measurement set-up and procedure are described in \cite{NaLuZhBeArBeCo17,Zhang2008a}. All other features like spatial dimensions of reconstruction volume, temporal sampling or the sub-sampling pattern are exactly the same as for the simulated data (cf. Section \ref{sec:trainOnVessel_clean}).

In-vivo data has different characteristics that are not perfectly represented by the training on synthetic data and hence a direct application of the trained network does not lead to satisfactory results, as illustrated by comparing it to a TV reconstruction in Figure \ref{fig:realData_cleanNet}. 
In particular, we see that the network has not learned to effectively threshold the noise-like artefacts in the low absorption regions i.e. regions with low concentration of chromophores. To train our approach to remove these features we simulated the effect of the low absorbing background as a Gaussian random field with short spatial correlation length, clipped the negative parts, scaled it to maximal value $0.1$ and added it to each segmented volume $x_{true}$ where ever the intensity of $x_{true}$ did not exceed $0.1$ (i.e., the maximum intensity of $x_{true}$ stays $1$).
%The phantoms with added background $x_{back}$ 
{The synthetic CT volumes with the added background} 
were then used for the data generation, i.e. $y^i_{back}=A x_{back}^i+\varepsilon$, whereas the clean volumes $x_{true}$ are used as reference for the training. Here $\varepsilon$ is again chosen (see Section \ref{sec:trainOnVessel_clean}) such that the resulting measurement $y^i_{back}$ had a SNR of approximately 15.
{We expect the network trained on the modified pairs $\{y^i_{back},x_{true}^i\}_i$ to be capable of effectively removing the background.}
%With this modified training set $\{y^i_{back},x_{true}^i\}_i$ we trained the network to remove background structures.

Furthermore, since the expected contrast in the images is crucial
for the trained reconstruction procedure, we %have 
scaled the measurement as follows. First we computed the standard deviation of the measurement data for all simulated targets. Then we %have 
rescaled the sub-sampled real measured data to have a similar standard deviation. This rescaled data is then used for reconstructing with the DGD. The result after 5 iterations is shown in Figure \ref{fig:RealDataResult}.

{The results can be further improved performing}
%To improve the results one can perform 
a transfer training of the previously trained networks $G_{\theta_k}$. This %would require that we have some kind of 
{however requires some} reference reconstructions %available from a similar or the same system. 
{from the same or a similar system.}
We were able to perform such a transfer training with a set of 20 (fully sampled) measurements %from the same system 
of a human finger, wrist, and palm {from the same system}.
We %have 
then sub-sampled the data (fourfold) to obtain the training data $y_{real}$. %As reference reconstruction we have taken  
{As reference we took} weakly regularised TV reconstructions from the fully sampled data, %as desired outcome, denoted as 
$x_{TV}$. To update the DGD we have performed an additional \revisNew{10} epochs 
{of training on the pairs}
%on the training set 
$\{y_{real},x_{TV}\}$, with a {reduced} learning rate of \revisNew{$10^{-5}$.} {Such transfer training} takes only 90 minutes {to update the entire DGD}. %whole updated DGD, 
We denote the updated CNNs by ${G}_{\hat{\theta}_k}$ and the resulting outputs by $\hat{x}_k$. The effect of the updated DGD is shown in Figure \ref{fig:RealDataResult}. 

\revisNew{Additionally, for a full comparison we have performed an update training of the U-Net with the same parameters as above, i.e. 10 epochs and a reduced learning rate of $10^{-5}$. The update training of the network took only 20 minutes and the result is shown in Figure \ref{fig:RealDataResult}.}

\section{Discussion of results\label{sec:Discussion}}

The results shown in Figure \ref{fig:SimulationResult} and Figure \ref{fig:RealDataResult} suggest that the formulation of a gradient descent scheme as a CNN in each iteration does produce competitive results with a considerable reduction in iterations needed, as we will discuss in this section. Furthermore, it is robust in the transition to real measurement data, which is one of the most important aspects in inverse problems and image reconstruction.

During the reconstruction procedure, a major improvement is achieved in the first step, as shown in Figure \ref{fig:ProgressIterations}. After one iteration of the DGD the background is cleared and the contrast is mostly restored, but there are still a few noisy patches around the vessels visible. The difference image also indicates that there are still parts insufficiently recovered on the outer area close to the boundary; these are typical limited view artefacts. After the 5th iteration these artefacts are considerably reduced and the error inside the domain is mostly uniform.

In the following, we discuss some particular aspects in more detail.

\subsection{Quantitative analysis of simulated data}

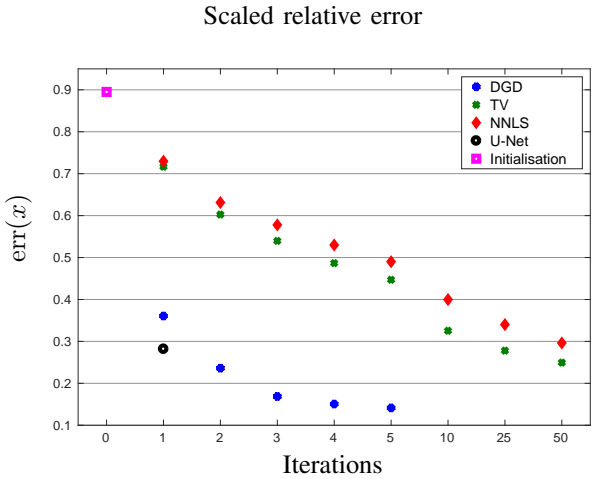
\begin{figure}[h!]
\centering
\begin{picture}(200,175)
\put(-20,0){\includegraphics[width=250 pt]{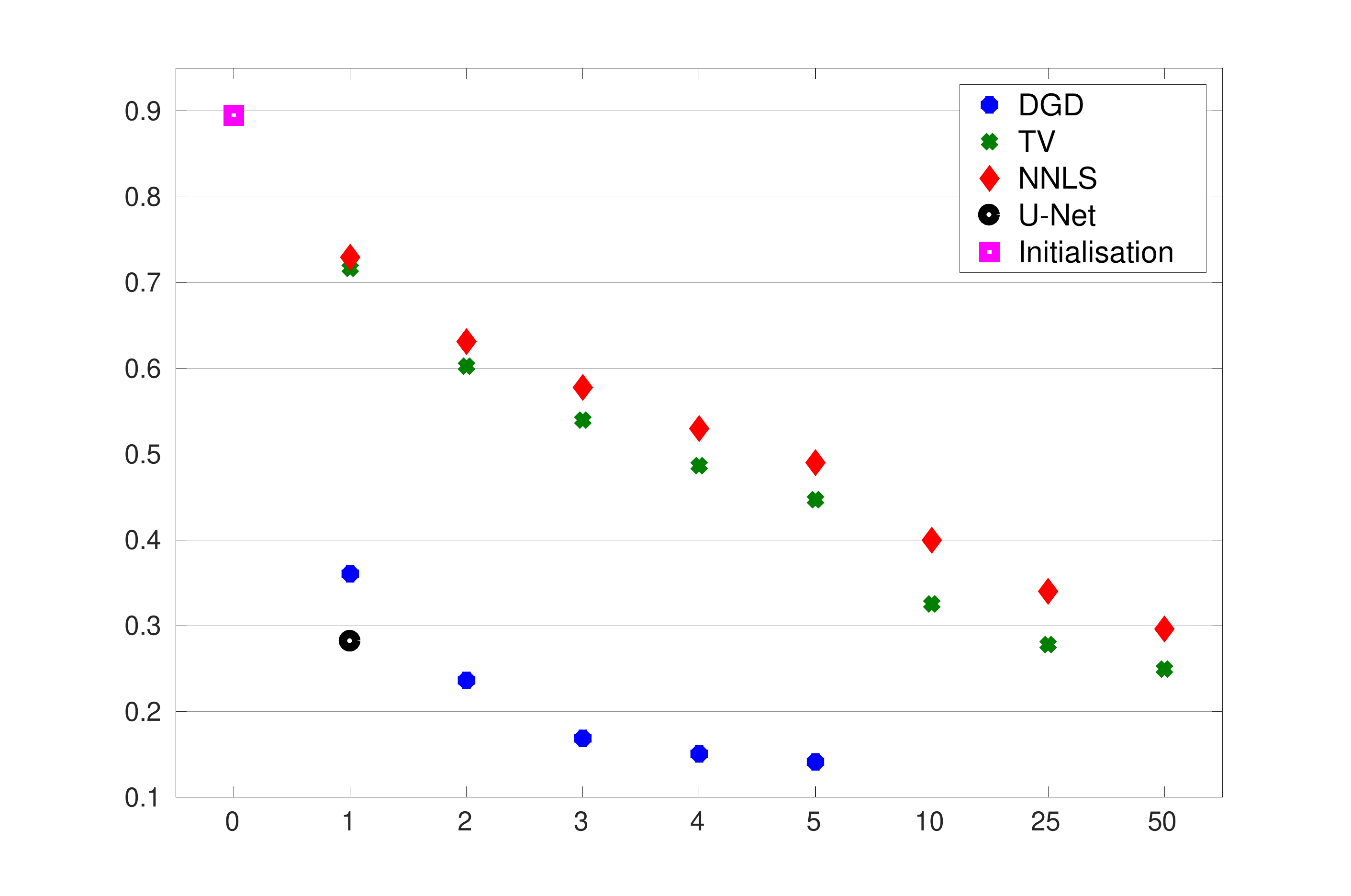}}
\put(60,170){Scaled relative error}
\put(90,0){Iterations}
\put(-15,80){\rotatebox{90}{$\mathrm{err}(x)$}}
\end{picture}
\caption{\label{fig:Convergence} \revision{Convergence plot of mean error for 16 samples from simulated test data}. The x-axis shows number of iterations (note the nonlinear scale). The y-axis denotes the unbiased relative $\ell^2$-errors \eqref{eqn:errMeas}. The parameter for TV has been chosen such that the best error is achieved for the given iterations.}
\end{figure}

For a quantitative evaluation of the performance we have computed the relative $\ell^2$-error for the simulated example shown in this study, see e.g. Figure \ref{fig:realData_cleanNet}. More precisely the reconstruction quality is evaluated using a scaled and unbiased relative error defined by
\begin{equation}\label{eqn:errMeas}
\mathrm{err}(x)=\min_{a,b}\frac{\|ax-x_{true}-b\|_2}{\|x_{true}\|_2},
\end{equation}
\revision{as suggested in \cite{JiMcFrUn17}. This unbiased error is used to not disadvantage TV and NNLS reconstructions in the comparison. While the networks know the absolute contrast from the training data, classical iterative methods often either need many iterations to recover it from the data or suffer from systematic contrast errors. Consequently, the {optimal} parameters for the reconstructions of DGD and U-Net are in most cases $a=1$ and $b=0$ and hence $\mathrm{err}$ reduces to the standard relative $\ell^2$-error. %Furthermore, only the relative contrast is of interest in PAT, since the reconstructions do not encode quantitative information. 
For a full comparison, we have computed the mean error for 16 test samples that were not included in the training set. We compare the two networks, U-net and DGD, with TV and NNLS reconstructions, as described in Section \ref{sec:PAT}, with the regularisation parameter for TV chosen such that $\mathrm{err}(x)$ is minimized.} The resulting errors are plotted in Figure \ref{fig:Convergence}. After one iteration U-Net achieves clearly the best result, but already with 2 iterations DGD achieves a smaller error down to a substantially smaller error after 5 iterations. TV and NNLS converge considerably slower, but achieve the U-Net quality after 50 iterations and will likely go lower.

The computational time is %governed 
{dominated} by the application of $A$ and its adjoint $A^*$\revision{. Computing either takes about 12 seconds on the Titan Xp GPU, see Table \ref{table:evalTimes} for the complete computation times for each reconstruction approach. Note however, that as our implementations involve communication overhead between Matlab and Python, theses timings give an indication for the methods' efficiency rather than an absolute comparison.}
Consequently, a reduction in iterations has a considerable impact on the total computation time. In this respect, the U-Net structure is clearly the cheapest with just one application to compute $x_0=A^*y$. Iterative algorithms require additionally two applications for each iterate to compute the gradient $\nabla d(y,Ax_k)=A^*(Ax_k-y)$. Thus, having similar results after 2 iterations with DGD and 50 iterations of TV, see Figure \ref{fig:Convergence}, leads to a prospective speed-up by 20 (including the initial reconstruction $x_0=A^*y$). \revision{We note that the computation time for U-Net can be considerably reduced by using a k-space method \cite{kostli2001temporal} for the initial reconstruction.}

\begin{table}[h] 
\centering
\scriptsize
  \caption{\revision{Evaluation times: including initialization $x_0=A^* y$ \revisNew{and communication overhead} for DGD and U-Net}}
    \begin{tabular}{l|c|c|c|c}
    %\hline
&  {\sc DGD} &{\sc TV} &{\sc NNLS} & {\sc U-Net}  \\
& 5 iterations & 5 iterations & 5 iterations & \\
 
    \hline
    \hline
    {\sc Time in sec.} & 184.19 &   165.72   & 147.7 & 19.75  \\
    %\hline
    %\hline
    \end{tabular}%
  \label{table:evalTimes}%
\end{table}%

\subsection{Comparison to post-processing by Deep Learning}\label{sec:compPostProc}
\begin{figure*}[t!]
\centering
\begin{picture}(500,320)
\put(0,170){\includegraphics[width=140 pt]{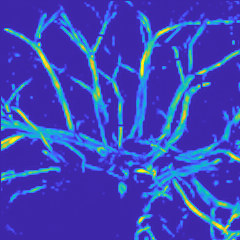}}
\put(170,170){\includegraphics[width=140 pt]{imagesSimData/SimData_Ex1_iter5_mIX.png}}
\put(340,170){\includegraphics[width=140 pt]{imagesSimData/SimData_Ex1_phantom_mIX.png}}
\put(50,320){U-Net}
\put(220,320){DGD $x_5$}
\put(390,320){Phantom}

\put(0,0){\fbox{\includegraphics[width=140 pt]{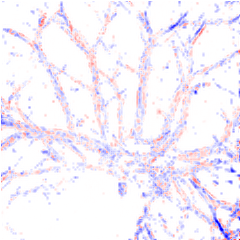}}}
\put(170,0){\fbox{\includegraphics[width=140 pt]{imagesSimData/SimData_Ex1_diff2True_iter5_mIX.png}}}

\put(30,150){Difference: U-Net}
\put(200,150){Difference: DGD $x_5$}

\put(340,93){\fbox{\includegraphics[width=140 pt]{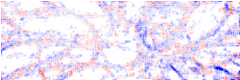}}}
\put(340,0){\fbox{\includegraphics[width=140 pt]{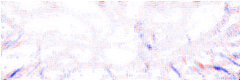}}}

\put(375,150){Difference: U-Net}
\put(374,60){Difference: DGD $x_5$}

\put(485,-3){{\includegraphics[height=146 pt]{images/blue2redHorzNumbers.png}}}
\put(485,167){{\includegraphics[height=146 pt]{images/ParulaHorzNumbers.png}}}

\end{picture}
\caption{\label{fig:simulation_CompareUnet} Comparison of reconstructions for a test image from the segmented CT data. 
Images are top-down maximum intensity projections and the difference images are on the same scale, with blue for a negative difference and red for positive.. Left: top and bottom shows the result by applying U-Net to the initialization $x_0$ and the difference to the phantom, maximal value of difference is $0.6012$. Middle: shows the result of the DGD after 5 iterations and the difference to the phantom, maximal value of difference is $0.4081$. Right bottom: difference images as side projections for the results of DGD and U-Net.}
\end{figure*}

First using a direct reconstruction and then applying a DNN to remove artifacts is a valid approach in many applications, especially if one is interested in fast and prospectively real-time reconstructions. This approach only needs an initial direct reconstruction and one application of the trained network. Especially for full-view data, this is a promising approach, but even in our limited-view case this approach proves to be quite powerful. 
A comparison of DGD and U-Net for simulated data is shown in Figure \ref{fig:simulation_CompareUnet} (top row). The resulting image is cleaned up and many vessels are properly reconstructed. Some smaller details are missing and can not be %reconstructed 
{recovered} from the initial reconstruction. The difference to the true target is also shown in Figure \ref{fig:simulation_CompareUnet} (bottom row). %Large differences are visible
{The differences are most pronounced} in the outer parts of the domain {as a consequence of} %connected to 
the limited view geometry.
In comparison the reconstruction by DGD has a much smaller 
overall {error}, but {this is} especially {true} in the center of the domain. The maximal  
{error} of the U-net reconstruction is $0.6012$ %(of maximal 1) 
(on the scale of $[0,1]$)
and of the DGD reconstruction $0.4081$. In conclusion we can say that the U-net architecture performs very well and is even capable of removing some limited-view artefacts, but is ultimately limited by the information contained in the initial reconstruction.

\subsection{In-vivo data}

Even though the results for simulated data are very impressive, applying the DGD trained on images with a clean background is not sufficient for real data as shown in Figure \ref{fig:realData_cleanNet}. {The reason is that} the algorithm %detects 
{interprets}
all structures in the data as important and enhances {them} equally. Adding a background to the {training data set} in order to teach the DGD thresholding those structures immensely improves the results and even fine details that were not visible before are {now} recovered after 5 iterations, as seen in Figure \ref{fig:RealDataResult}. \revision{Nevertheless, just an adjustment of the simulated data is not sufficient as can be seen from the quantitative measures in Table \ref{table:inVivoErrors}, computed with respect to the reference reconstruction from fully-sampled limited-view data}. Thus, further improvement can be achieved by an update of the DGD if one has a set of similar measurements from fully sampled data available. \revision{This update training has a considerable impact on the reconstruction quality as can been seen in Table \ref{table:inVivoErrors}.}
\revisNew{Both learned methods show excellent reconstruction quality after transfer training and are able to successfully remove the undesired background structures. In comparison to the iterative reconstruction with TV both learned methods achieve a higher PSNR and SSIM to the reference reconstruction from fully-sampled data. Noteworthy, the lowest unbiased relative $\ell^2$-error ($\mathrm{err}$), see \eqref{eqn:errMeas}, is achieved by the classical TV minimisation with an emphasis on the data term, this is likely due to the fact that the reference is a TV reconstruction from fully-sampled data.}

\begin{table}[h] 
\scriptsize
  \caption{\revisNew{Quantitative measures for in-vivo experiment: in comparison to reference TV reconstruction from fully-sampled limited-view data.}}
    \begin{tabular}{l|c|c|c|c}
     %\multicolumn{5}{c}{\sc Reconstructed Average Values}\\
    %\hline
&{\sc PSNR} &{\sc SSIM} & {\sc err}& {\sc rel. $\ell^2$-error}  \\
    \hline
    \hline
    {\sc DGD $x_5$} & 32.93&   0.723   & 0.76 & 1.54  \\
    {\sc Updated DGD $\hat{x}_5$}  & \bf{41.40} &  \bf{0.945}  &  0.56 & \bf{0.58}   \\
    {\sc U-Net}  & 40.81 &  0.933  &  0.62 & 0.62   \\
    {\sc TV sub-samp., $\lambda=5\cdot 10^{-5}$}  & 38.05  & 0.912  &  \bf{0.52} & 0.86   \\
    {\sc TV sub-samp., $\lambda=10^{-4}$}  & 37.68 &  0.902  &  0.58  & 0.89 \\    
    \hline
    \hline
    \end{tabular}%
  \label{table:inVivoErrors}%
\end{table}%

\subsection{\revision{Generalisation and robustness}}
\revision{Deep Learning approaches are especially powerful in a fixed measurement protocol and consistent targets, as illustrated for the simulated test data. The big question is how robust these networks are with respect to perturbations of measurement procedures or targets. First experiments indicate that the iterative network allows for small perturbations in the forward operator such as varying sub-sampling patterns (of same sub-sampling rate) or deviations in sound speed, as well as slightly varying noise level in the data. However, each variation will lead to slight deterioration of reconstruction quality. In contrast, the one step approach by U-Net was found much more sensitive to variations. In particular, we have found that a change in sampling pattern leads to a mean (for 16 samples) deterioration in $\mathrm{err}$ by $0.5\%$ for DGD and U-Net by $5\%$ for the simulated test data. We think that this is due to the fact, that the gradient in each iteration encodes the model variations and as such small perturbations are corrected in the iterative network. If larger changes in the measurement protocol are expected, it is recommended to either retrain the network or perform an update training, as has been done for the in-vivo data.}

\begin{figure*}[h]
\centering
\begin{picture}(510,140)
\put(00,0){\includegraphics[width=120 pt]{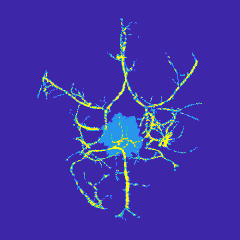}}
\put(125,0){\includegraphics[width=120 pt]{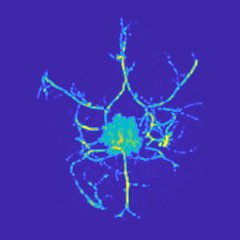}}
\put(250,0){\includegraphics[width=120 pt]{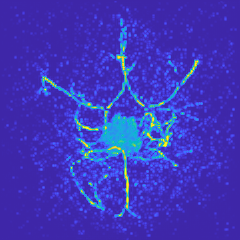}}
\put(375,0){\includegraphics[width=120 pt]{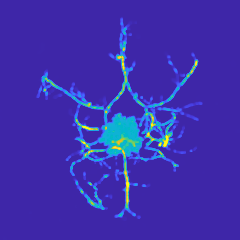}}
\put(497,-2){{\includegraphics[height=125 pt]{images/ParulaHorzNumbers.png}}}
\put(45,125){Phantom}
\put(172,125){DGD $x_5$}
\put(300,125){U-Net}
\put(430,125){TV}
\end{picture}
\caption{\label{fig:tumorPhantom} \revision{Reconstruction of a tumor phantom with features that are not included in the training data. DGD and U-Net reconstructions are done with the networks trained on the segmented vessel phantoms. The TV reconstruction is computed with 20 iterations and a regularisation parameter $\lambda=10^{-4}$. Reconstruction errors with the unbiased $\mathrm{err}$ are: DGD 0.4925, U-Net 0.6584, TV 0.4749.}}
\end{figure*}

\revision{Furthermore, the iterative method seems to be more robust with respect to structural differences between the target and the training set. This is illustrated in Figure \ref{fig:tumorPhantom}, where we have tested the networks trained on the clean segmented vessels on a tumor phantom \cite{ArBeBeCoHuLuOgZh16}. With 5 iterations we achieve a similar $\mathrm{err}$ as TV after 20 iterations. As it can be seen, the network does reproduce vessels with similar characteristic as in the training set, this might be due to the learned prior-like filters. Whereas the U-Net reconstruction does not perform well with the new image structures.}

\section{Conclusions}\label{sec:Conclusions}
In limited-view, sub-sampled photoacoustic tomography it is essential to incorporate the physical model into the reconstruction procedure to reduce artefacts \revisNew{with an appropriate regularisation strategy. Here we considered three possible strategies: i) iterative total variation, ii) backprojection followed by a learned denoiser, iii) learned iterative reconstruction. In terms of image quality and robustness to perturbations in the model i) and iii) were superior to ii). Method ii) was fastest at the cost of inferior image quality and flexibility. Method iii) was considerably faster than i).}
{Thus, we believe that learned iterative reconstructions are a realistic technique for 3D PAT. The choice between learned post-processing versus learned iterative reconstruction is a matter of speed versus quality.}

{This study is particularly focused on method iii) and} we have shown that incorporating the physical model as the gradient of the data fit and learning an iterative algorithm consisting of several convolutional neural networks leads to a superior reconstruction quality with a considerable speed-up compared to classical, and well established, iterative reconstruction schemes. With minor modifications we were able to apply the learned algorithm to experimental in-vivo data of a human wrist and obtained far more detailed reconstructions from sub-sampled data than by TV minimisation of the same data. 

Additionally, we have investigated method ii) that consists of post-processing a fast and basic direct reconstruction with a CNN, in particular we implemented an architecture introduced as U-Net that has been proven to work well on medical images. In our study this approach shows promise to produce a fast and good initial reconstruction, but since many features are not present in simple direct reconstructions, for limited-view, sub-sampled data, this approach is limited by the quality of the initial reconstruction. Even though certain features can not be recovered, post-processing with Deep Learning is promising for applications where 
low latency is more important than a best quality reconstruction,
%best reconstruction quality is not needed and rather a low latency is of importance, 
such as navigational tasks during surgery.
\revision{Furthermore, our study suggests that iterative networks are more robust with respect to changes in the measurement setup or imaged target.}

{As inherent in all learning approaches, the limitation of the proposed method is dictated by the quality of the training data and the possibility to perform an update training.}
In future research we will consider combing the U-Net architecture with a model based approach. For instance by replacing the CNNs representing one iteration in our \emph{deep gradient descent} with a U-Net like structure. For high resolution 3D imaging this would need computational resources exceeding a local workstation. Consequently, if the computational %cost becomes more feasible, 
{resources are available}
including the forward operator in the training will likely improve results even further.

% use section* for acknowledgment
\section*{Acknowledgment}
We gratefully acknowledge the support of NVIDIA Corporation with the donation of the Titan Xp GPU used for this research. This study was supported in part by funding from the European Union's Horizon 2020 research and innovation programme H2020 ICT 2016-2017 under grant agreement No 732411 and is an initiative of the Photonics Public Private Partnership. \\
AH acknowledges support from the Wellcome-EPSRC project NS/A000027/1, "Image-Guided Intrauterine Minimally Invasive Fetal Diagnosis and Therapy" (GiftSurg). FL acknowledges financial support from EPSRC project EP/K009745/1 "Dynamic High Resolution Photoacoustic Tomography System" and of the Netherlands Organization for Scientific Research (NWO), project nr. 613.009.106/2383. JA acknowledges Swedish Foundation of Strategic Research grants AM13-0049 and ID14-0055 as well as support from Elekta.

%\vfill

% Can be used to pull up biographies so that the bottom of the last one
% is flush with the other column.
%\enlargethispage{-5in}

\bibliographystyle{unsrt}
\bibliography{literature,PAT,PATCS}
%\bibliography{literature,PAT}

% that's all folks
\end{document}